\documentclass[letterpaper]{article} 
\usepackage{aaai2026}  
\usepackage{times}  
\usepackage{helvet}  
\usepackage{courier}  
\usepackage[hyphens]{url}  
\usepackage{graphicx} 
\usepackage{natbib}  
\usepackage{caption} 
\usepackage{algorithm}
\usepackage{algorithmic}
\usepackage{array}
\usepackage{booktabs}
\usepackage{amsmath,amsfonts}
\usepackage{enumitem} 

\usepackage{newfloat}
\usepackage{listings}
\DeclareCaptionStyle{ruled}{labelfont=normalfont,labelsep=colon,strut=off} 
\floatstyle{ruled}
\newfloat{listing}{tb}{lst}{}
\floatname{listing}{Listing}
\title{Edge-Centric Relational Reasoning for 3D Scene Graph Prediction}
\author{
    Yanni Ma\textsuperscript{\rm 1,2},
    Hao Liu\textsuperscript{\rm 3,4},
    Yulan Guo\thanks{Corresponding author.}\textsuperscript{\rm 1},
    Theo Gevers\textsuperscript{\rm 2},
    Martin R. Oswald\textsuperscript{\rm 2}
}
\affiliations{
    \textsuperscript{\rm 1}School of Electronics and Communication Engineering, Sun Yat-sen University, Shenzhen, China \\
    \textsuperscript{\rm 2}Computer Vision Research Group, University of Amsterdam, Netherlands \\
    \textsuperscript{\rm 3}Key Lab of Spatial-temporal Big Data Analysis and Application of Natural Resources in Megacities, \\ Ministry of Natural Resources, East China Normal University (ECNU), Shanghai, China\\
    \textsuperscript{\rm 4} Key Laboratory of Geographic Information Science (Ministry of Education), ECNU, Shanghai, China\\
    mayn3@mail2.sysu.edu.cn,
    hliu@geoai.ecnu.edu.cn,
    guoyulan@sysu.edu.cn,
    th.gevers@uva.nl,
    m.r.oswald@uva.nl

}

\begin{document}

\maketitle

\begin{abstract}

3D scene graph prediction aims to abstract complex 3D environments into structured graphs consisting of objects and their pairwise relationships. Existing approaches typically adopt object-centric graph neural networks, where relation edge features are iteratively updated by aggregating messages from connected object nodes. However, this design inherently restricts relation representations to pairwise object context, making it difficult to capture high-order relational dependencies that are essential for accurate relation prediction. To address this limitation, we propose a \textbf{L}ink-guided \textbf{E}dge-centric relational reasoning framework with \textbf{O}bject-aware fusion, namely \textbf{LEO}, which enables progressive reasoning from relation-level context to object-level understanding. Specifically, LEO first predicts potential links between object pairs to suppress irrelevant edges, and then transforms the original scene graph into a line graph where each relation is treated as a node. A line graph neural network is applied to perform edge-centric relational reasoning to capture inter-relation context. The enriched relation features are subsequently integrated into the original object-centric graph to enhance object-level reasoning and improve relation prediction. Our framework is model-agnostic and can be integrated with any existing object-centric method. Experiments on the 3DSSG dataset with two competitive baselines show consistent improvements, highlighting the effectiveness of our edge-to-object reasoning paradigm.  

\end{abstract}


\section{Introduction}

3D scene graph prediction (SGP) is a fundamental yet challenging task in 3D scene understanding, which aims to represent complex environments as structured graphs by recognizing objects and modeling their relationships. 
Unlike conventional 3D scene understanding tasks, such as 3D object detection \cite{CenterTube,AnchorPoint, ao2022you} and semantic segmentation \cite{GCR,liu2020semantic,dangSeg}, which mainly focus on identifying individual objects, 3D SGP goes a step further by capturing the rich semantic and spatial relations among them.
In a 3D scene graph, nodes correspond to object instances, and edges encode pairwise relations, such as ``behind'', or ``attached to''. This representation captures both the spatial arrangement and semantic interactions of objects, providing a compact and interpretable abstraction for high-level reasoning. Consequently, 3D SGP has been increasingly adopted in downstream tasks such as 3D visual grounding \cite{scanrefer, referit3d}, visual question answering \cite{scanqa,clevr3d,3dvqa-tcsvt} and embodied AI \cite{embodiedAI, Physcene}.

\begin{figure}[t]
\centering
\includegraphics[width=1\columnwidth]{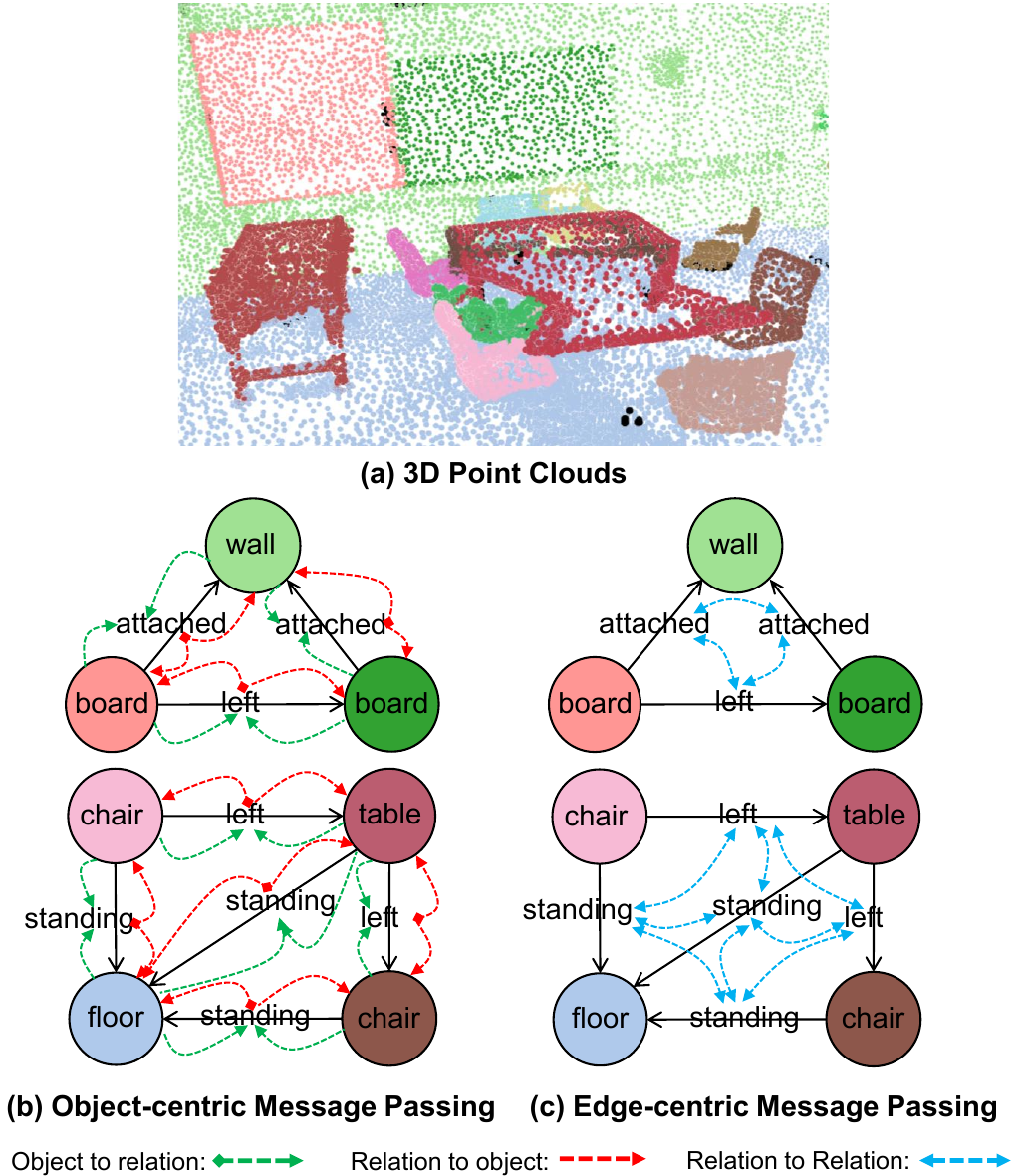} 
\caption{ Object-centric message passing vs. Edge-centric message passing. }
\label{novelty}
\end{figure}

To model such structured representations, existing methods widely adopt graph neural networks (GNNs), with various architectural designs proposed for enhancing relational reasoning. For instance, SGPN \cite{3DSSG} and KISGP \cite{KISGP} employ fully-connected GNNs to propagate messages between object nodes and their associated relations. 3DSMKA \cite{3DSMKA} introduces a hierarchical graph to capture multi-level spatial and semantic structures, while 3DHetSGP \cite{3DHetSGP} constructs a heterogeneous graph to explicitly account for the diversity of relation types. These models have demonstrated strong performance on the 3DSSG benchmark, highlighting the importance of GNN-based relational reasoning. However, most existing methods follow an object-centric paradigm, where messages are iteratively propagated between objects and relations. Although this bidirectional design allows modeling of local object–relation interactions, it inherently restricts the relational reasoning to pairwise context, as each relation is updated independently based on its connected objects. As a result, these methods struggle to capture higher-order relational dependencies such as predicate co-occurrence or contextual consistency among relations. 

This limitation is particularly evident in indoor scenes \cite{3DSSG}, where object semantics and geometric layouts alone are often insufficient for accurate relation prediction. For example, semantic ambiguity is a common issue: relations like ``hanging on'', ``attached to'', and ``supported by'' often share similar spatial cues and are easily misclassified. Moreover, spatial relations are often interdependent, \textit{e.g.}, an object \textit{to the left of} another may also be \textit{behind} it, depending on global scene layout. Without modeling such relation dependencies, object-centric approaches often produce inaccurate or inconsistent scene graphs.

To address these limitations, we propose a \textbf{L}ink-guided \textbf{E}dge-centric relational reasoning framework with \textbf{O}bject-aware fusion, namely \textbf{LEO}, which enables progressive reasoning from relation-level context to object-level understanding. Specifically, LEO first predicts soft link weights for each edge in the original scene graph to modulate the message passing strength. Then, we transform the original graph into a line graph, where each node represents a relation and two nodes are connected if the associated relations share a common object. A line graph neural network (LineGNN) is applied to perform edge-centric relational reasoning, enabling each relation to aggregate contextual information from other semantically related relations. Finally, the enhanced relation features are then integrated back into the original object-centric graph to improve object-level reasoning and relation prediction. LEO is model-agnostic and can be seamlessly integrated into existing object-centric approaches. Our main contributions are summarized as: 

\begin{itemize}

\item We propose \textbf{LEO}, a link-guide edge-to-object reasoning framework that performs progressive relational reasoning from edge-centric to object-centric representations for accurate and robust 3D scene graph prediction.

\item We introduce a line graph formulation for 3D scene graphs, enabling explicit relation-level reasoning beyond object pairs. To the best of our knowledge, this is the first work to reformulate 3D scene graphs as line graphs for relation-centric reasoning.

\item We design a link prediction module that assigns soft weights to object pairs, modulating relation strengths and suppressing noisy or irrelevant message passing.

\item Extensive experiments on the 3DSSG dataset show that our LEO consistently improves two strong baselines, demonstrating the effectiveness and generalizability of modeling relation-level dependencies. 

\end{itemize}

\section{Related Works} 
\paragraph{3D scene graph prediction in point cloud.}
Recent advances in 3D scene understanding have extended the paradigm of scene graph generation from 2D images to 3D point clouds, which aims to jointly infer object categories and their semantic relationships directly from point cloud data. As the pioneering work, SGPN \cite{3DSSG} introduces the first large-scale dataset 3DSSG for this task, and proposes a baseline framework that integrates PointNet \cite{Pointnet} for point-wise feature encoding with an object-centric GNN for joint reasoning over nodes and edges. To better capture contextual dependencies among objects and relations, SGFN \cite{SGFN} introduces a feature-wise attention mechanism within the GNN message passing process, while SGGpoint \cite{EdgeGCN} enhances relation reasoning through a twinning attention module designed for edge-oriented updates. KISGP \cite{KISGP} further incorporates category-level priors via a graph auto-encoder, enabling knowledge-guided message propagation, whereas 3DSMKA\cite{3DSMKA} leverages both external knowledge bases and spatial priors to construct a hierarchical graph that encodes multiscale semantics and geometry. VL-SAT \cite{VL_SAT} explores vision-language supervision by aligning CLIP-based features \cite{CLIP} with object and relation embeddings. More recently, 3DHetSGP \cite{3DHetSGP} proposes a heterogeneous graph learning framework that explicitly models the diversity of relation types and mitigates the long-tail distribution of predicates in 3D scene graphs. However, these methods follow an object-centric paradigm, which inherently restricts relation features to pairwise object context, leading to less accurate and robust relation predictions. In contrast, our approach explicitly models the dependencies among relations by reformulating the scene graph into a line graph. This enables edge-centric relational reasoning that captures higher-order context beyond individual object pairs.

\paragraph{Line graph neural network.}
Line graph \cite{LineGraph} is a classical edge-centric structure derived from an original graph, where each node in the line graph corresponds to an edge in the original graph, and two nodes are connected if their associated edges share a common object. Building upon this concept, Line graph neural network (LineGNN) treat edges as nodes and perform message passing between them, enabling relational reasoning at the edge level. Such models have been successfully applied in diverse domains, including molecular property prediction \cite{ESA}, network neuroscience \cite{neuroscience}, traffic forecasting \cite{traffic_fore}, and community detection \cite{community}, where modeling interactions among edges is essential. These successes highlight the potential of edge-centric reasoning, motivating our adoption of LineGNN to enhance relation-level reasoning in 3D scene graphs.


\section{Preliminary}
\subsubsection{Line Graph.}
Given an original graph $ \mathcal {G} = (\mathcal{V}, \mathcal{E})$, its line graph $ \mathcal L( \mathcal {G}) = (\mathcal {V_L}, \mathcal {E_L})$ is a derived graph where each node corresponds to  an edge in $\mathcal{G}$.  Two nodes in $\mathcal{L}(\mathcal{G})$  are adjacent if and only if their corresponding edges in $\mathcal{G}$ share a common node.   
For a fully connected original graph  $ \mathcal {G} = (\mathcal{V}, \mathcal{E})$, the number of nodes in its line graph equals the number of the edges in $\mathcal{G}$, i.e., $|\mathcal{V_L}| = |\mathcal{E}|$, and the number of edges in the line graph becomes $|\mathcal{V_L}|(|\mathcal{V}|-2)$. Such dense connectivity in $\mathcal{L}(\mathcal{G})$ enables comprehensive modeling of relation-level context. 

\begin{figure}[t]
\centering
\includegraphics[width=0.9\columnwidth]{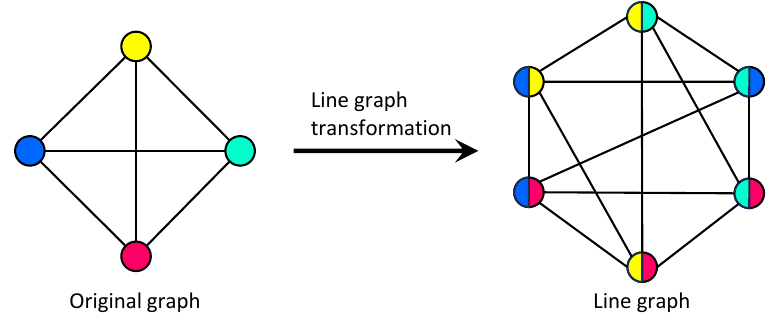} 
\caption{ Line Graph Transformation. }
\label{linegraph}
\end{figure}

\begin{figure*}[!t]
\centering
\includegraphics[width=0.93\textwidth]{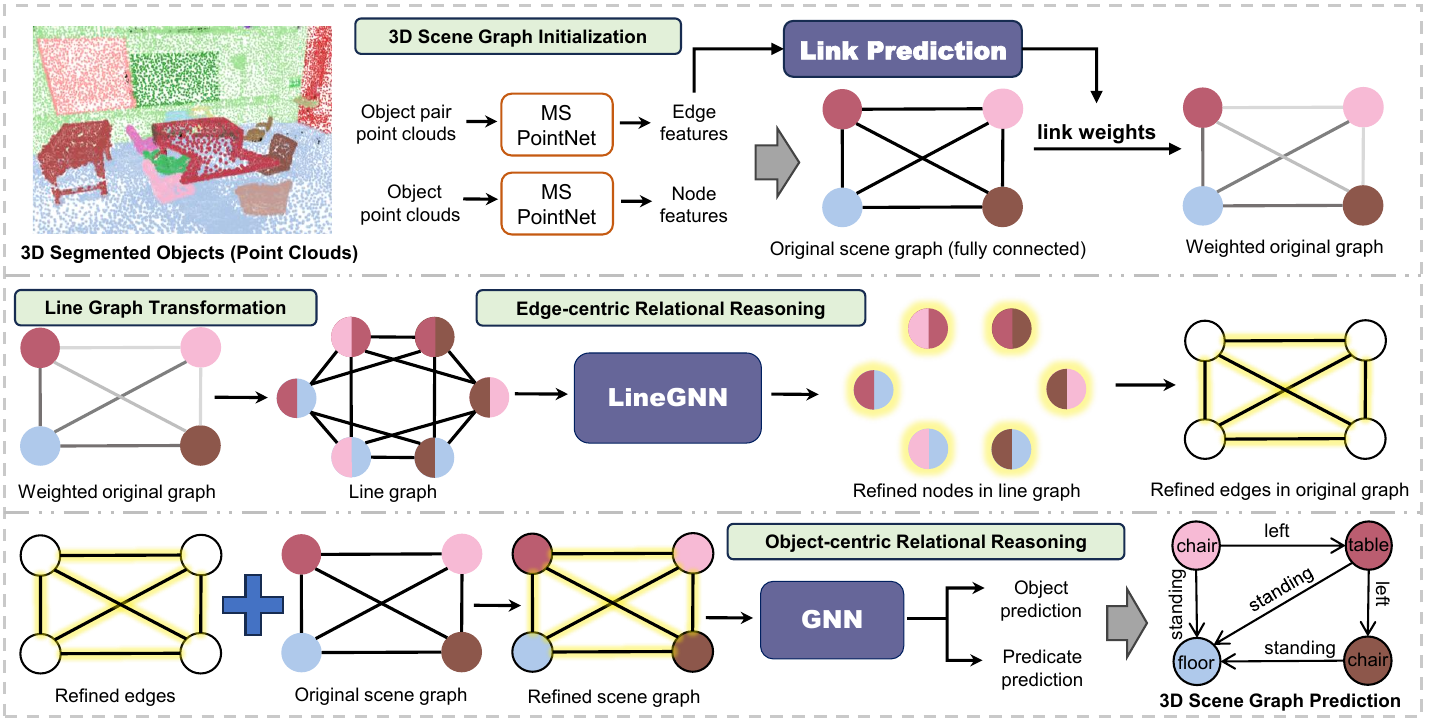}
\caption{ The overview of our LEO framework. It consists of three stages: (a) Link Prediction assigns soft link weights to object pairs in the original scene graph to modulate relation strengths for subsequent reasoning; 
(b) Edge-centric Relational Reasoning transforms the weighted original graph into a line graph and applies  LineGNN to capture relation-level context and refine relation features; (c) Object-centric Relational Reasoning integrates the refined relations into the original graph for final object and predicate prediction.
}
\label{fig_1}

\end{figure*}

\section{Methods}
\subsection{Overview}

Given a segmented point cloud $\mathcal{P} \in \mathbb{R}^{N \times 3}$ with $N$ 3D points and a set of $K$   class-agnostic instances masks $\mathcal {M}=\{\mathbf {M}_1,\dots, \mathbf {M}_K\}$, the objective of 3D scene graph prediction is to construct a structured graph $\mathcal{G} = (\mathcal{V}, \mathcal{E})$ of the 3D scene by identifying object instances $\mathcal{V}$ and predicting relations  predicates $\mathcal{E}$ between them. 

As illustrated in Fig.~\ref{fig_1},  we propose Link-guided Edge-to-Object Reasoning network (LEO),  a unified framework  that performs sequential edge-centric and object-centric relational reasoning. 
Specifically, it consists of three key components: 
(1) Link Prediction assigns soft link weights to object pairs in the original scene graph, modulating relation strengths for subsequent reasoning.
(2) Edge-centric Relational Reasoning transforms a line graph from original graph and applies a Line Graph Neural Network (LineGNN) to model inter-relation context.
(3) Object-centric Relational Reasoning integrates the enhanced relation features into the original scene graph for final object and predicate classification.
This design allows LEO to effectively capture both relation-level dependencies and object-level context, leading to more accurate and consistent 3D scene graph predictions.

\subsection{3D Scene Graph Initialization}
We first build a fully connected primitive graph  $G=(V,E) $ to represent all possible pairwise relations among object instances. Following our baseline models KISGP \cite{KISGP} and 3DHetSGP \cite{3DHetSGP}, the initial object and edge features are encoded and pretrained separately. The object features $f_n$ are extracted from segmented point sets using a multi-scale PointNet encoder \cite{Pointnet}. For edge features $f_e$, we compute the difference between subject and object feature pairs, followed by an MLP projection. If object class labels are available (e.g., in the PredCls setting), the object features are substituted with the object label embeddings.

\subsection{Link Prediction}
To mitigate the redundancy from dense connections in the line graph, we introduce a link prediction module 
to predict the likelihood of  correlation between object pairs. The predicted scores are subsequently assigned as soft weights to the edges in the original graph, thereby modulating the relation strengths and suppressing irrelevant message passing during relational reasoning. 

Given a set of objects with their initialized features $f_i$ and spatial attributes of bounding boxes $b_i$,  we first compute the geometric embedding: 
\begin{equation}
\label{geom_feature}
g_i =\phi_b ([b_i, c_i, l_i, w_i, h_i,V_i]),
\end{equation}
where $\phi_b(\cdot)$ is the non-linear transformation. $b_i$ and $c_i$ denote the box corners and centroid, while  $l_i$, $w_i$, $h_i$,$V_i$ represent the box length, width, height,  volume, respectively. 
For each pair $(i,j)$, we construct the link features $f_{ij}^{link}$ by concatenating the  differences of the initialized object features  $f_i$, $f_j$ and geometric embeddings $g_i$, $g_j$:

\begin{equation}
\label{pair_feature}
f_{ij}^{link} = \phi_p ([(f_i-f_j) \| (g_i-g_j)]),
\end{equation}
where $\phi_p(\cdot)$ is the non-linear transformation for concatenated features. 
The link features $f_{ij}^{link}$ are then classified  into two link categories (link and non-link) using a linear classifier $\phi_l(\cdot)$, followed by a 2-way softmax that yields the confidence scores for both categories: 
\begin{equation}
\label{link_classify}
s_{ij}^{link} = \mathrm{softmax} (\phi_l f_{ij}^{link})
\end{equation}
We take the probability of the positive link as the final link confidence score, which  serves as the soft link weight  $s_{ij}^{link}$  between objects $i$ and $j$ to guide relational reasoning.  

\subsection{Edge-centric Relational Reasoning}
\subsubsection{Weighted Original Graph}
Based on the soft link weights predicted by the link prediction module,  we convert the original graph $\mathcal {G}$ to a weighted  graph by multiplying the link weights with all edge features.
\begin{equation}
\label{weight_Edge}
\tilde f_{ij} = s_{ij}^{link} \cdot f_{ij}
\end{equation}
This weighted graph encodes the importance of each   relation for modulating the message passing in LineGNN.

\subsubsection{Line Graph Transformation}
To enable edge-centric relational reasoning, we transform the original graph $\mathcal{G}= (\mathcal{V}, \mathcal{E})$ into a line graph $\mathcal{L}(\mathcal{G}) = (\mathcal{V}', \mathcal{E}')$. 
In this line graph, each node $e_{ij} \in \mathcal{V}'$ represents a directed edge between object nodes $(o_i, o_j)$ in the original graph. There exists an edge between two line nodes $e_{ij}$ and $e_{ik}$ if they share the same object node $o_i$ in $\mathcal{G}$, indicating a contextual relation dependency. Formally:
\begin{equation}
\label{Line_node}
\mathcal{V}'=  \{ e_{ij} | (o_i, o_j) \in \mathcal{E} \},
\end{equation}
\begin{equation}
\label{Line_edge}
\mathcal{E}'=  \{ (e_{ij}, e_{ik}) | e_{ij} \cap e_{ik}= o_i \in V \}.
\end{equation}

\subsubsection{{Line Graph Neural Network}}
To model interactions among relation edges, we apply a Line Graph Neural Network (LineGNN) over our line graph $\mathcal{L}(\mathcal{G}) = (\mathcal{V}', \mathcal{E}')$ to propagate and aggregate relational context features. In LineGNN, message passing is performed between adjacent relation nodes based on the connectivity defined in $\mathcal{E}'$.
Specifically, let $h^{(l)}_{ij}$ denote the hidden state of relation node $ e_{ij}$ at layer $l$.  
At each layer, the feature of $e_{ij}$ is updated by aggregating messages from its neighboring nodes in $\mathcal{V}'$.

\noindent \textbf{a) Incoming Message. }
Given a relation node $e_{ij}$, its incoming message $ \boldsymbol{m}^{(l)}_{ij}$  is computed by aggregating the features of its neighboring relation nodes $e_{ik} \in \mathcal{N}(i)$ as:
\begin{equation}
\label{MP_linenodes}
    \boldsymbol{m}^{(l)}_{ij} =  \mathrm {LN} \left(\sum_{e_{ik} \in \mathcal{N}_{e_{ij}}} \alpha^{(l)}_{ij \rightarrow ik}  \phi_e(\boldsymbol{h}_{ik}^{(l)})\right), 
\end{equation}
where  $\phi_e(\cdot)$ is a non-linear transformation, $ \mathcal{N}_{e_{ij}} $ is the neighboring relations set of relation node $e_{ij}$, and $\text{LN}(\cdot)$ denotes layer normalization. The attention score  $\alpha^{(l)}_{ij \rightarrow ik}$ ensures that more relevant neighbors contribute more to the message,  computed as:
\begin{equation}
\alpha^{(l)}_{ij \rightarrow ik} = \text{softmax}(\phi_{\text{att}}([h^{(l)}_{ij} \| h^{(l)}_{ik}] )),
\end{equation}
where $[\cdot \| \cdot]$ denotes the concatenation operation, $\phi_{\text{att}}(\cdot)$ is a non-linear transformation, the \text{softmax($\cdot$)} is computed over all neighbor relations  $e_{ik} \in \mathcal{N}_{e_{ij}}$.

\noindent \textbf{b) Edge-centric Updating. }
Each relation node then updates its hidden state via a gated recurrent unit (GRU):
\begin{equation}
\label{MP1-h_nodes}
 \boldsymbol{h}^{(l+1)}_{ij} = \mathrm {GRU}(\boldsymbol{h}_{ij}^{(l)} ,\boldsymbol{m}^{(l)}_{ij}).
\end{equation}

The relation features $\tilde{h}^{(l+1)}_{ij}$ output from the last LineGNN layer encode rich contextual dependencies among relations and are subsequently utilized as the starting edge features in the first layer of primitive message passing for object-centric reasoning.

\subsection{Object-centric Relational Reasoning}
After enriching the relation features via the LineGNN module, we propagate the updated relation representations to the primitive graph to perform object-centric reasoning.  
The initial hidden state in this primitive graph is defined as:
\begin{equation}
\boldsymbol{h}_{i}^{(0)} = f_i ,     \boldsymbol{h}_{ij}^{(0)} = \tilde{h}^{(l+1)}_{ij},
\end{equation}
where $\tilde{h}^{(l+1)}_{ij}$ denotes the output from the lineGNN module.  

Each object node receives messages from its connected edges to update its hidden state accordingly. Meanwhile, edge features are also updated based on the hidden states of their associated object nodes.  
\begin{align}
\label{MP-nodes-edges}
    \boldsymbol{h}_{i}^{(l+1)} &=  \text{GRU}(\boldsymbol{h}_{i}^{(l)}, \boldsymbol{m}^{(l)}_i)  \nonumber  \\
    &= \text{GRU}\Big(\boldsymbol{h}_{i}^{(l)} , \mathrm {LN} (\sum_{j \in N_{i*}} \phi_e(\boldsymbol{h}_{ij}^{l}))\Big), \\
\label{MP2}
 \boldsymbol{h}_{ij}^{(l+1)} &=  \text{GRU}(\boldsymbol{h}_{ij}^{(l)}, \boldsymbol{m}^{(l)}_{ij}) \nonumber \\ 
 &= \text{GRU}\Big(\boldsymbol{h}_{ij}^{(l)} ,  \mathrm {LN}(\phi_n(\boldsymbol{h}_{i}^{l}) + \phi_n(\boldsymbol{h}_{j}^{l}))\Big).
\end{align}

\subsection{Scene Graph Prediction and Training Objective }
To predict the 3D scene graph, we classify the object features $f_n$ and edge features $f_e$ obtained from primitive graph message passing into object and predicate categories.
\begin{equation}
\label{object-predictor}
\boldsymbol{s}_n= \mathrm{softmax}\big( \phi_{obj}(\boldsymbol{f}_n)\big),
\end{equation}
\begin{equation}
\label{pred-predictor}
\boldsymbol{s}_{e}= \mathrm{softmax}\big(\phi_{pred}(\boldsymbol{f}_{e})\big).
\end{equation}
Our scene graph prediction involves object classification, predicate classification and link classification. Therefore the overall training objective of  is defined as:
\begin{align}
\mathcal L_{total} = \mathcal L_{obj} + \mathcal L_{pred} +  \mathcal L_{link}, 
\end{align}
where the $\mathcal L_{link}$ refers to link loss of  link prediction module. 
We use focal loss for all loss components.

\section{Experiments}
\subsection{Experimental Settings}
\subsubsection{Dataset.}
The 3DSSG dataset provides annotated 3D semantic scene graphs built upon the 3RScan dataset, encompassing 1,482 scans from 478 indoor environments. It contains approximately 48k object nodes and 544k relation edges, capturing rich spatial and semantic relationships among objects. Following prior work, we split the dataset into 3,852 sub-scenes for training and 548 for testing, where each sub-scene contains 4 to 9 objects. 160 object categories and 26 predicate categories are adopted  for training and evaluation, consistent with the RIO27 annotation scheme.

\subsubsection{Metrics. }
We evaluate our model on two standard tasks of 3D scene graph prediction: (1) Predicate Classification (PredCls), where the model only predicts the predicate category for each object pair given the ground truth object labels; and (2) Scene Graph Classification (SGCls), which requires the model to predict both object categories and the relationships between objects.
For both tasks, we adopt the following metrics: top-k recall (R@k), no-graph-constraint top-k  recall (ngcR@k), and mean recall (mR@k). Specifically , R@k measures the propotion of ground truth triplets recalled among the top-k highest-scoring predictions, with constraint that each subject-object pair is assigned with a single predicate. In contrast, ngcR@k allows multiple predicates per object pair without this constraint.
The mR@k metric computes the average recall across all predicate categories, providing a more balanced evaluation and better captures performance under long-tail distributions.

\begin{table*}[!t]
    \centering
    \resizebox{0.98\textwidth}{!}{
    \begin{tabular}{l|*{3}{>{\centering\arraybackslash}p{1.3cm}}| *{3}{>{\centering\arraybackslash}p{1.45cm}} | *{3}{>{\centering\arraybackslash}p{1.3cm}}}
        \toprule
        Model & R@20  & R@50  & R@100 & ngcR@20  & ngcR@50  &  ngcR@100  & mR@20  & mR@50  &  mR@100 \\
        \midrule
        Co-Occurrence \cite{KISGP} & 34.7 & 47.4 & 47.9 &  35.1 &  55.6 &  70.6 & 33.8 & 47.4 & 47.9 \\
        KERN \cite{KERN}   & 46.8 & 55.7 & 56.5 & 48.3 & 64.8 & 77.2 & 18.8 & 25.6 & 26.5 \\
        Schemata \cite{Schemata}  & 48.7 & 58.2 & 59.1& 49.6 & 67.1 & 80.2 & 35.2 & 42.6 & 43.3 \\
        SGPN \cite{3DSSG}   & 51.9 & 58.0 & 58.5 & 54.5 & 70.1 & 82.4 & 32.1 & 38.4 & 38.9\\
        SGFN* \cite{SGFN}  & 54.5 & 61.0 & 61.5 & 61.4 &  80.1 &  90.0 & 30.5  & 36.4  & 36.6\\
        VL-SAT* \cite{VL_SAT} & 58.3  &   65.2  &   65.8 & 66.2  & 85.9  &  93.9 & 40.4 &  47.4 &  47.7 \\
        3DSMKA \cite{3DSMKA} & - &     68.3  &    69.5 & - & 79.8  & 89.6  & - &  66.5  & 66.9  \\
        SGFormer* \cite{SGFormer} & 53.6 & 59.8 &  60.2 & 56.2 & 71.7 & 83.5 & 37.2 &  43.1 & 43.4 \\
        KISGP \cite{KISGP}  & 59.3 & 65.0 &  65.3 & 62.2 & 78.4 & 88.3 & 56.6 & 63.5 & 63.8 \\
        3DHetSGP* \cite{3DHetSGP} &    61.8 &   65.9 &   65.9 &   70.3 &   88.9 &   94.6  &   63.7 &   68.1 &  68.2 \\
        \midrule
        KISGP* \cite{KISGP}  & 59.9 &  65.1 &  65.4 & 63.1 & 79.1 & 88.6 & 57.1 & 61.9 & 62.0\\
        KISGP+LEO  (Ours)  &  \bf 61.5 & \bf  66.8  &  \bf 66.9  &  \bf 64.4  &  \bf 80.3 &  \bf 89.6 &  \bf 59.0  &  \bf 64.7  &   \bf 64.8 \\
        \midrule
        3DHetSGP*  &    61.8 &   65.9 &   65.9 &   70.3 &   88.9 &   94.6  &   63.7 &   68.1 &  68.2 \\
        3DHetSGP+LEO (Ours)  & \bf 62.9 &  65.8 &  65.8 &  \bf 73.3 &  \bf 90.1  &  \bf 95.1 &  \bf 65.8 &  \bf  68.5 & \bf  68.9 \\
        \bottomrule
    \end{tabular}}
    \caption{Quantitative results of the evaluated methods in PredCls tasks.  Models marked with * indicate reproduced results.}
    \label{tab:table1}
\end{table*}

\begin{table*}[!t]
\centering
\resizebox{0.98\textwidth}{!}{
    \begin{tabular}{l|*{3}{>{\centering\arraybackslash}p{1.3cm}}| *{3}{>{\centering\arraybackslash}p{1.45cm}} | *{3}{>{\centering\arraybackslash}p{1.3cm}}}

\toprule
Model & R@20  & R@50  & R@100 & ngcR@20  & ngcR@50  &  ngcR@100  & mR@20  & mR@50  &  mR@100 \\
\midrule 
Co-Occurrence \cite{KISGP} &  14.8 &  19.7 &  19.9 &  14.1 &  20.2 &  25.8 &  8.8 &  12.7 &  12.9 \\
KERN \cite{KERN}  &  20.3 &  22.4 &  22.7 &  20.8 &  24.7 &  27.6 &  9.5 &  11.5 &  11.9 \\
Schemata \cite{Schemata} &  27.4 &  29.2 &  29.4 &  28.8 &  33.5 &  36.3 &  23.8 &  27.0 &  27.2\\
SGPN \cite{3DSSG}  &  27.0 &  28.8 &  29.0 &  28.2 &  32.6 &  35.3  &  19.7 &  22.6 &  23.1 \\
SGFN* \cite{SGFN} &  27.2 &  28.9 &  28.9 &  30.0 &  34.6 &  36.9 &  18.2 &  20.8 &  20.9   \\
VL-SAT* \cite{VL_SAT} &   29.5  &  31.0  & 31.2  &  32.8 &   37.7  & 39.8 &    27.5 &    29.9 &  30.0 \\
3DSMKA \cite{3DSMKA} & - &    31.5  &   31.6  & - & 35.4   & 37.7   & - &  30.3 & 30.6 \\
SGFormer* \cite{SGFormer} & 28.9  &  30.7 &  30.8 &  30.2 & 34.7 & 37.2 &  24.1 &  26.2 &26.2\\
KISGP \cite{KISGP} &  28.5 &  30.0 &  30.1 &  29.8 &  34.3 &  37.0 &  24.4 &  28.6 &  28.8 \\
3DHetSGP* \cite{3DHetSGP} &   28.5 &   29.8 &   29.9 &   31.2 &  36.4 & 38.7  &   27.3 &     29.4 &   29.5  \\
\midrule
KISGP* \cite{KISGP} &   28.8 &   30.5 &   30.6 &  30.3 &  34.7 &  37.6 &  25.1 &  28.1 &  28.3\\
KISGP+LEO (Ours)  &  \bf 29.5  & \bf 30.9  & \bf 31.0   &  \bf 32.5  & \bf 37.4  & \bf 39.5  & \bf 27.5 & \bf 30.4 & \bf 30.4  \\
\midrule
3DHetSGP* \cite{3DHetSGP} &  \bf 28.5 &   29.8 &   29.9 &   31.2 &  36.4 &  \bf 38.7  &   27.3 &     29.4 &   29.5  \\
3DHetSGP+LEO (Ours)   &  28.4 & \bf 30.1 & \bf  30.2 & \bf  31.9 & \bf 36.5 & 38.5 & \bf  29.6 & \bf  32.3 & \bf  32.3 \\
\bottomrule
\end{tabular}}
\caption{\label{tab:table2}Quantitative results of the evaluated methods in SGCls tasks. Models marked with * indicate reproduced results. }
\end{table*}

\begin{figure*}[!t]
\centering
\resizebox{0.98\textwidth}{!}{\includegraphics[width=4.5in]{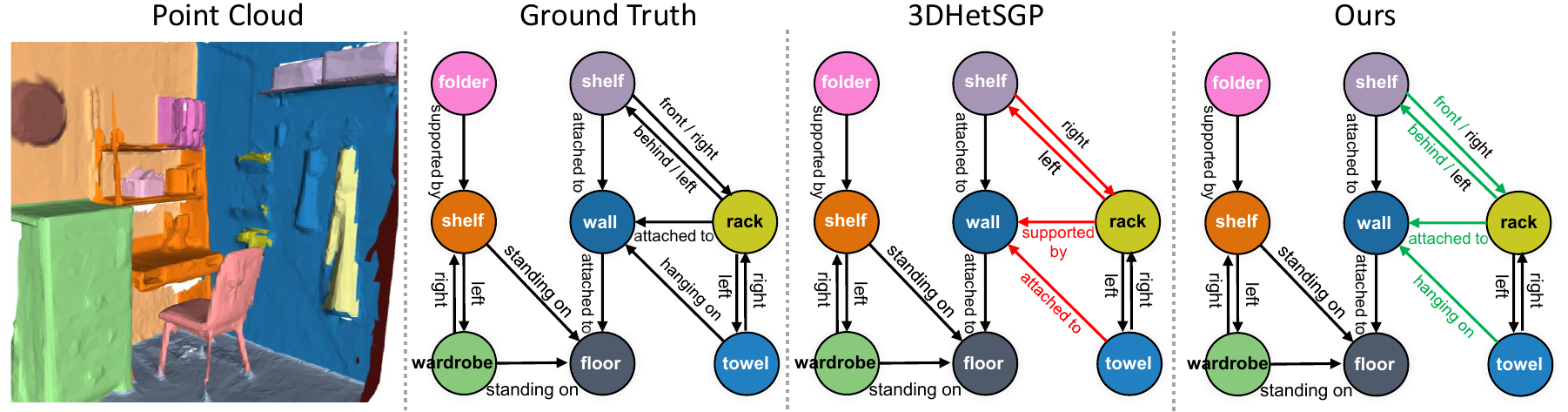}} \\
\caption{Qualitative results of our model and baseline 3DHetSGP \cite{3DHetSGP} on ngcR@20.  Red arrows indicate incorrectly predicted relationships. Green arrows indicate relationships that are missed or misclassified by 3DHetSGP but correctly predicted by ours. }
\label{Q1}
\end{figure*}

\subsubsection{Implementation details.}
Our model is implemented in PyTorch and trained on a single NVIDIA RTX TITAN GPU. We follow the same training configuration as the baseline models KISGP  \cite{KISGP}  and 3DHetSGP \cite{3DHetSGP}  for fair comparison. Specifically, we use the ADAM optimizer with an initial learning rate of $0.0001$ and  weight decay of $0.0001$.  The learning rate decays by a factor of 0.7 every 10 epochs (minimum 1e-8).
Following  KISGP and 3DHetSGP, we  pretrain the object and predicate encoders  based on multi-scale PointNet. 
The LineGNN is inserted before the original object-centric GNN. In particular in 3DHetSGP, which employs three GNN branches for type subgraphs, we also insert LineGNN before GNN independently in each branch.
Our experiment run $40$ epochs for link prediction, and $50$ epochs for relational reasoning stage to predict the final 3D scene graph.

\begin{table*}[ht]
\centering
\resizebox{0.95\textwidth}{!}{
\begin{tabular}{l|*{3}{>{\centering\arraybackslash}p{1.3cm}}| *{3}{>{\centering\arraybackslash}p{1.4cm}} | *{3}{>{\centering\arraybackslash}p{1.3cm}}}
\toprule
PredCls  & R@20  & R@50  & R@100 & ngcR@20 & ngcR@50 & ngcR@100 & mR@20  & mR@50  &  mR@100 \\[0.5ex] 
\midrule
3DHetSGP (Baseline)  &  61.8 &   65.9 &    65.9 &   70.3 &   88.9 &   94.6  &    63.7 &   68.1 &  68.2 \\
\midrule
LineGNN (FC)      &  60.4 & 64.3  & 64.4   &  68.6  & 88.5 & 94.6 & 60.0  &  66.0  &  66.1 \\
LineGNN (LP) &  62.9 &  65.8 &  65.8 &   73.3 &   90.1  &   95.1 &   65.8 &    68.5 &   68.9 \\
LineGNN (GT)      &  63.2 &  65.1 &  65.1  &  77.8 &  94.3 & 97.7 &  66.6 &  69.0 &  69.3  \\
\midrule
\midrule
SGCls  & R@20  & R@50  & R@100 & ngcR@20 & ngcR@50 & ngcR@100 & mR@20  & mR@50  &  mR@100 \\[0.5ex] 
\midrule
3DHetSGP (Baseline) &   28.5 &   29.8 &   29.9 &   31.2 &  36.4 & 38.7  &   27.3 &     29.4 &   29.5  \\
\midrule
LineGNN (FC)      &  29.5  & 30.9  & 31.0   &  32.5  & 37.4  & 39.5  &  27.5 &  30.4 &  30.4 \\
LineGNN (LP) &   28.4 &  30.1 &    30.2 &    31.9 &  36.5 & 38.5 &  29.6 &    32.3 &   32.3 \\
LineGNN (GT)      &  32.2 &  32.7 &  32.8  &  37.2 &  41.5 & 42.4 & 31.6 &  33.1 & 33.1  \\
\bottomrule
\end{tabular}}
\caption{\label{tab:AB_LP}Ablation study on the effects of different link weights for LineGNN.  ``FC'' denote the fully connected original graph, ``LP'' denotes the edges in original graph processed with predicted soft link weights, ``GT'' denotes the edges in original graph filtered by link ground truth. }  
\end{table*}

\subsection{Quantitative results}
Tables \ref{tab:table1} and \ref{tab:table2} report the results on the PredCls and SGCls tasks, respectively.
To evaluate the effectiveness and generalizability of the proposed Le framework, we integrate it into two representative baselines: KISGP and 3DHetSGP, which adopt object-centric graph reasoning architectures and have achieved leading performance on the 3DSSG benchmark. We further compare our method with a set of advanced 3DSGP models in point clouds, including SGPN \cite{3DSSG}, \cite{SGFN}, SGFormer \cite{SGFormer}, VL-SAT \cite{VL_SAT}, and 3DSMKA \cite{3DSMKA}.
Methods marked with $\star$  are reproduced by us using their open-source code, and all results are evaluated using the KISGP evaluation protocol to ensure consistency. 

\subsubsection{LEO on KISGP. }
When  integrated into KISGP, LEO brings consistent  improvements. On the PredCls task (Table \ref{tab:table1}), we observe gains of +2.2\%, +2.0\%, and +0.6\% at ngcR@20/50/100, respectively. The improvement at ngcR@20 (from 62.2\% to 64.4\%)  indicates that more correct relations are ranked among the top 20 predictions, suggesting enhanced accuracy in high-confidence relation predictions.
The mean recall also increases from 63.5\% to 64.7\% at mR@50, and from 63.8\% to 64.8\% at mR@100, indicating more balanced coverage of predicate categories.
On the SGCls task (Table \ref{tab:table2}), LEO  also increases ngcR@50 from 34.7\% to 37.4\%, and mR@50 from 28.1\% to 30.4\%. 
This suggests that LEO enhances 3DSGP performance under both settings where object information is provided as ground-truth labels or inferred from visual features.

\subsubsection{LEO on 3DHetSGP. }
We further integrate  LEO  to a stronger baseline, 3DHetSGP, and observe consistent improvements. On the PredCls task,  LEO improves R@20 from 61.8\% to 62.9\%, ngcR@20 from 70.3\% to 73.3\%, and mR@20 from 63.7\% to 65.8\%. 
The concurrent improvements indicate that the proposed edge-centric relational reasoning improves the accuracy and ranking of predicted relations, particularly among high-confidence predictions.
On the SGCls task, we also observe modest gains, particularly in mean Recall@20/50/100, indicating that the model achieves more balanced predictions acro ss predicate categories, including those with lower frequency.

\subsubsection{Overall Comparison. }
Compared to both KISGP and 3DHetSGP, integrating the edge-centric relational reasoning  consistently improves performance across various metrics. 
The gains are more evident on KISGP, which uses a relatively simple GNN-GRU architecture without explicit modeling of relation-level interactions. 
On 3DHetSGP, which adopts a heterogeneous graph structure with multiple type edges, the improvements are smaller but consistent. 
These results indicate that edge-centric relational reasoning provides consistent benefits across scene graph models with varying GNN architectures, including both standard and heterogeneous GNN-based frameworks, suggesting its general applicability to 3D scene graph prediction.

\subsection{Qualitative results}
Fig \ref{Q1} presents a qualitative comparison between our method and the baseline 3DHetSGP on ngcR@20. 
For the object pair \textit{towel} and \textit{wall}, the ground truth relation is \textit{hanging on}. 3DHetSGP identifies the relation predicate  as \textit{attached to}, likely due to its limited capacity to distinguish among visually similar predicates. In contrast, our method successfully predicts \textit{hanging on}, demonstrating edge-centric relational reasoning can handle semantic confusion by leveraging relational context.
For the pair \textit{shelf} and \textit{rack}, the ground truth includes two directional spatial relations:  \textit{behind},  \textit{left} and  \textit{front},  \textit{right}. 
3DHetSGP fails to predict \textit{behind} and \textit{front} among the top 20 predicted relations. 
Our method accurately predicts both relations, suggesting that it more effectively captures the inter-relation between spatial predicates through edge-centric relational reasoning. More qualitative results are provided in the supplementary material.

\subsection{Ablation study}
\subsubsection{Effect of link prediction for LineGNN. }
Table \ref{tab:AB_LP} reports the effects of different link weights in the original graph on LineGNN. We consider three settings: using a fully connected link weights (FC), applying a link prediction module to obtain confident link weights (LP), and using ground-truth as link weights (GT). Compared to FC, LP  consistently improves performance across both PredCls and SGCls, especially on ngcR@k and mR@k, showing that suppressing noisy edges before the line graph transformation benefits edge-centric relational reasoning. The GT variant yields the best results, serving as a reference to illustrate the performance upper bound under ideal link guidance.
Note that the GT setting is included solely for analysis and is not considered part of our proposed method. 
These results confirm that the effectiveness of LineGNN  relies significantly  on the quality of links in the original graph, and that link prediction provides a practical solution to enhance relational context without requiring additional annotations.

\subsubsection{Ablation Study on LineGNN Depth.}
We evaluate the impact of LineGNN depth by varying the number of layers from 1 to 7. As shown in Table \ref{tab:GNNdepth}. The results show that the best performance is achieved with 5 layers in terms of mean Recall (mR@20: 65.80, mR@50: 68.49, mR@100: 68.92). 
Based on these results, 5 layers is adopted as the default setting for LineGNN.

\begin{table}[t]
  \centering
  \resizebox{\columnwidth}{!}{
    \begin{tabular}{c|ccc|ccc}
    \toprule
      &  \multicolumn{3}{c|}{\textit{Ngc Recall}} & \multicolumn{3}{c}{\textit{mean Recall}}    \\ 
    \midrule
        layers &  ngcR@20 & ngcR@50 & ngcR@100 & mR@20 & mR@50 & mR@100\\
    \midrule
         1 &  72.30  & 89.91 & 95.11  & 64.17 & 66.82 & 67.14 \\
         2 & 73.11  & \bf 90.59 & \bf 95.62 & 64.94 & 68.41 & 68.78 \\
         3 & 73.40 & 90.33 & 95.20 & 62.98 & 66.89 & 67.03 \\
         4 & 73.27 & 90.22 & 95.54 & 63.79 & 67.03 & 67.39 \\
         5 & 73.30 & 90.15 & 95.14& \bf 65.80 & \bf 68.49 & \bf 68.92 \\
         6 & \bf 73.48 & \bf 90.59 & 95.50 & 65.22 & 68.09 & 68.56 \\
         7 & 73.28 & 90.57 & 95.40 & 63.96 & 67.30 &  67.68 \\
    \bottomrule
    \end{tabular}}
    \caption{Ablation on layer numbers of LineGNN. }
  \label{tab:GNNdepth}
\end{table}

\subsubsection{Ablation on LineGNN Integration Strategies. }
We evaluate four strategies for integrating LineGNN into the object-centric framework: inserting it before the object-centric GNNs (\textbf{Pre}), after (\textbf{Post}), mixing the object-centric GNN with LineGNN  (\textbf{Mix}), and concatenating outputs from parallel branches (\textbf{Para}). As shown in Table \ref{tab:LGNNplace}, the \textbf{Pre} strategy consistently outperforms the others, achieving the highest ngcR@20 (73.30) and mR@100 (68.92), indicating that injecting relational context early benefits subsequent object-centric reasoning. In contrast, \textbf{Post} performs the worst, especially in mR@20 (61.60), suggesting that reasoning over objects first may limit the ability to capture relational dependencies. \textbf{Mix} and \textbf{Para} yield moderate performance, but still fall behind \textbf{Pre}. These results demonstrate that early integration of LineGNN is the most effective strategy for enhancing 3D scene graph prediction.
\begin{table}[ht]
  \centering
  \resizebox{\columnwidth}{!}{
    \begin{tabular}{l|ccc|ccc}
    \toprule
      &  \multicolumn{3}{c|}{\textit{Ngc Recall}} & \multicolumn{3}{c}{\textit{mean Recall}}    \\ 
    \midrule
        - &  ngcR@20 & ngcR@50 & ngcR@100 & mR@20 & mR@50 & mR@100\\
    \midrule
         Pre & \bf 73.30 & 90.15 & 95.14 & \bf 65.80 & \bf 68.49 & \bf 68.92 \\
         Post &  72.58 & 90.07 & 94.91 &  61.60 & 66.52 & 66.53  \\
         Mix & 72.57 & 89.98 &  95.22 & 62.22 & 67.06 & 67.18 \\
         Para & 72.34 & \bf 90.35 & \bf 95.69 &  62.43 & 67.11 & 67.17 \\
    \bottomrule
    \end{tabular}}
    \caption{Ablation on LineGNN integration strategies. }
  \label{tab:LGNNplace}
\end{table}

\section{Conclusion}
In this paper, we presented LEO, a unified framework for 3D scene graph prediction that performs sequential reasoning from edge-centric to object-centric paradigms. To address the limitations of conventional object-centric GNNs in capturing rich inter-relation context, we first construct a line graph where each relation is represented as a node, and apply edge-centric relational reasoning using a Line Graph Neural Network. The enhanced relation features are then integrated back into the original scene graph to facilitate object-aware reasoning. Extensive experiments on the 3DSSG dataset demonstrate that LEO consistently improves the performance of two strong baselines, validating the effectiveness of modeling relation-level dependencies through our edge-to-object reasoning paradigm.

\section{Acknowledgements} This work was partially supported by the National Natural Science Foundation of China (No. U20A20185, 62372491), the Guangdong Basic and Applied Basic Research Foundation (2022B1515020103, 2023B1515120087), the Shenzhen Science and Technology Program (No. RCYX20200714114641140). This work was also supported by the Oversea Study Program of Guangzhou Elite Project. Part of this work was conducted at the Computer Vision Group, University of Amsterdam, whose support is gratefully acknowledged.

\bibliography{main}

\clearpage

\newpage

\section{A. Additional Experiments}
\noindent  \textbf{A1. Efficiency Comparisons.} 
To evaluate the computational efficiency of our proposed LineGNN module, 
we compare its inference time, GPU memory usage, and FLOPs against the corresponding baselines, namely KISGSP and 3DHetSGP, as summarized in Table~\ref{tab:Computation}. 
Specifically, KISGSP+LineGNN increases inference time from 20.52\,ms to 23.50\,ms  and GPU memory usage from 0.70\,GB to 0.95\,GB, while maintaining a comparable FLOP count (1.61\,G vs. 1.82\,G). 
For 3DHetSGP, incorporating LineGNN results in an increase from 17.18\,ms to 25.99\,ms in inference time and from 1.03\,GB to 1.90\,GB in memory, with FLOPs remaining in a similar range (1.22\,G vs. 1.85\,G). 
These results confirm that the proposed edge-centric reasoning achieves higher relational accuracy with only limited computational overhead, showing good scalability scalability in both runtime and memory.

\begin{table}[ht]
  \caption{Efficiency comparison of inference time, GPU memory usage, and FLOPs.} 
  \label{tab:Computation}
  \centering
    \resizebox{0.48\textwidth}{!}{
    \begin{tabular}{l|ccc}
        \toprule
        \textbf{Model} & \textbf{Inf. Time (ms)} & \textbf{FLOPs (G)} & \textbf{GPU Mem. (GB)}  \\
        \midrule
        KISGP & 20.52 & 1.61 & 0.70 \\
        KISGP+LineGNN & 23.50 & 1.82 & 0.95 \\
        \midrule
        3DHetSGP & 17.18 & 1.22 & 1.03 \\
        3DHetSGP+LineGNN & 25.99 & 1.85 &  1.90 \\
        \bottomrule
    \end{tabular}}
\end{table}

\noindent  \textbf {A2. Effect of link prediction without LineGNN.}
To further disentangle the effect of link modulation from relation-level aggregation, we additionally evaluated a no-LineGNN + link-prediction variant, where the predicted link weights were applied to filter the original graph, but the LineGNN module was removed. As shown in the new results (Fig. / Table X, “no-LineGNN + LP”), this variant achieves moderate improvements over the fully connected baseline, confirming that the link-prediction module alone can effectively suppress noisy or spurious edges. However, its performance remains clearly below that of the full LineGNN model (Table \ref{tab:AB_LPLEO}), indicating that relation-level aggregation on the constructed line graph is indispensable for capturing higher-order contextual dependencies between predicates.
Together, these results clarify the complementary roles of the two components:
(1) link prediction refines the original graph topology by emphasizing reliable object pairs, while
(2) LineGNN subsequently exploits these refined connections to reason over inter-relation dependencies.
This additional ablation therefore isolates the gain from link modulation and justifies the necessity of the LineGNN for comprehensive edge-centric reasoning.

\begin{table*}[ht]
\centering
\resizebox{0.95\textwidth}{!}{
\begin{tabular}{l|*{3}{>{\centering\arraybackslash}p{1.3cm}}| *{3}{>{\centering\arraybackslash}p{1.4cm}} | *{3}{>{\centering\arraybackslash}p{1.3cm}}}
\toprule
PredCls  & R@20  & R@50  & R@100 & ngcR@20 & ngcR@50 & ngcR@100 & mR@20  & mR@50  &  mR@100 \\[0.5ex] 
\midrule
3DHetSGP*(Baseline)  &  61.8 &   65.9 &    65.9 &   70.3 &   88.9 &   94.6  &    63.7 &   68.1 &  68.2 \\
\midrule
LineGNN (FC)      &  60.4 & 64.3  & 64.4   &  68.6  & 88.5 & 94.6 & 60.0  &  66.0  &  66.1 \\
LineGNN (LP) &  62.9 &  65.8 &  65.8 &   73.3 &   90.1  &   95.1 &   65.8 &    68.5 &   68.9 \\
-no-LineGNN +LP &  62.3 &  65.4 &  65.6 &   73.2 &   90.5  &   95.0 &   64.4  &  67.7 &   68.3 \\
LineGNN (GT)      &  63.2 &  65.1 &  65.1  &  77.8 &  94.3 & 97.7 &  66.6 &  69.0 &  69.3  \\
\midrule
\midrule
SGCls  & R@20  & R@50  & R@100 & ngcR@20 & ngcR@50 & ngcR@100 & mR@20  & mR@50  &  mR@100 \\[0.5ex] 
\midrule
3DHetSGP*(Baseline) &   28.5 &   29.8 &   29.9 &   31.2 &  36.4 & 38.7  &   27.3 &     29.4 &   29.5  \\
\midrule
LineGNN (FC)      &  29.5  & 30.9  & 31.0   &  32.5  & 37.4  & 39.5  &  27.5 &  30.4 &  30.4 \\
-no-LineGNN +LP &   28.0 &  29.5 &  29.5 &   31.6 &  35.9 & 37.9 &  29.2 &    31.1 &   31.2\\
LineGNN (LP) &   28.4 &  30.1 &    30.2 &    31.9 &  36.5 & 38.5 &  29.6 &    32.3 &   32.3 \\
LineGNN (GT)      &  32.2 &  32.7 &  32.8  &  37.2 &  41.5 & 42.4 & 31.6 &  33.1 & 33.1  \\
\bottomrule
\end{tabular}}
\caption{\label{tab:AB_LPLEO}Ablation study on the effects of different link weights for LineGNN.  `FC' denote the fully connected original graph, `LP' denotes the edges in original graph are processed with predicted soft link weights, `no-LineGNN + LP' applies the predicted link weights while removing the LineGNN module. `GT' denotes the edges in original graph are filtered by link ground truth. }
\end{table*}

\section{B. More Qualitative results}
\begin{figure*}[t]
\centering

\resizebox{0.98\textwidth}{!}{\includegraphics[width=4.5in]{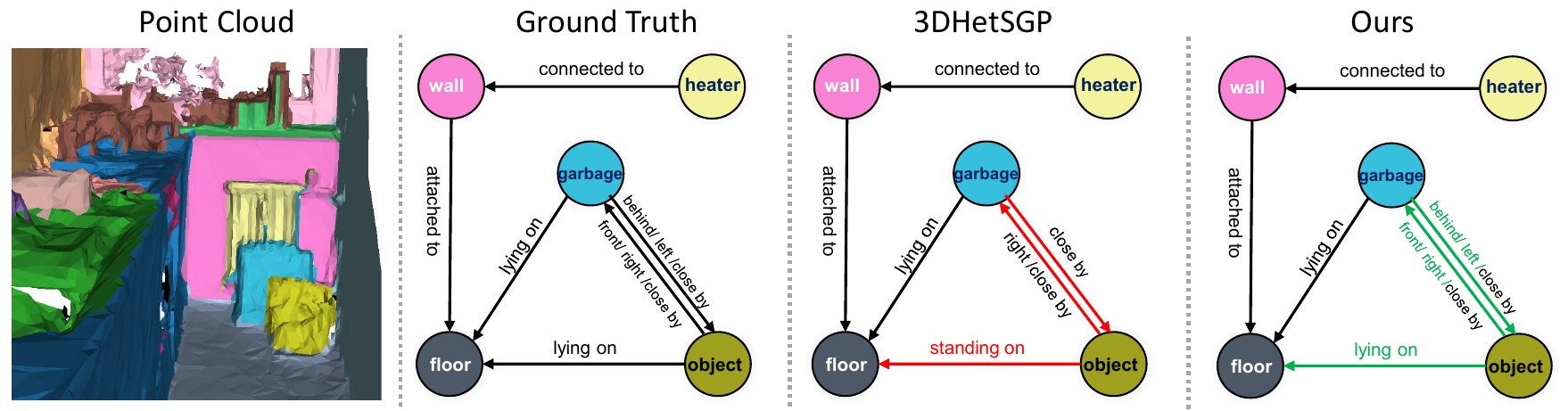}} \\
\resizebox{0.98\textwidth}{!}{\includegraphics[width=4.5in]{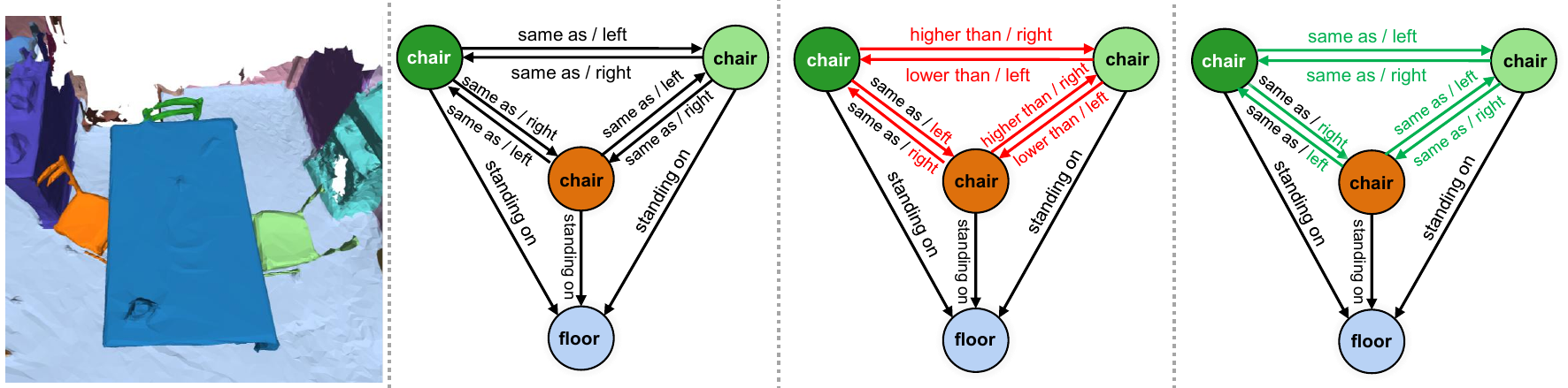}} \\
\resizebox{0.98\textwidth}{!}{\includegraphics[width=4.5in]{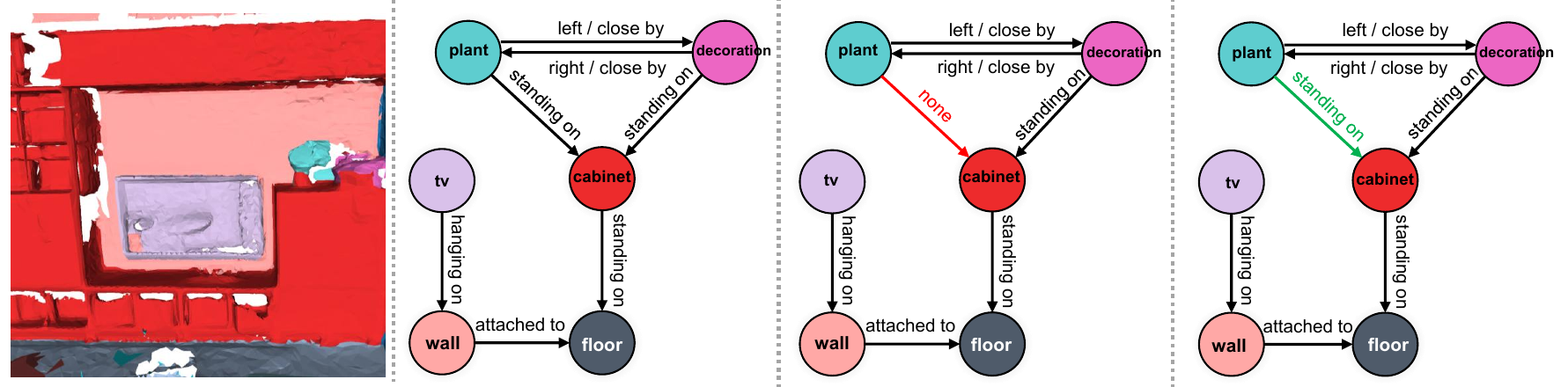}}%

\caption{Qualitative results of our model and baseline 3DHetSGP on 3DSSG dataset under ngcR@20.  Red arrows indicate incorrectly predicted relationships. Green arrows indicate relationships that are missed or misclassified by 3DHetSGP but correctly predicted by ours.}
\label{Q2}
\end{figure*}

\noindent  \textbf {B1. Qualitative Analysis on 3DSSG dataset.}
Figure \ref{Q2} presents additional qualitative comparisons on the 3DSSG dataset, evaluted under the ngcR@20. We visualize the predicted scene graphs from our method and the baseline 3DHetSGP  across three indoor scenes. In each case, we visualize the predicted scene graphs alongside the ground truth.

In the first example, 
3DHetSGP misclassifies the \textit{lying on} relation between  \textit{object} and  \textit{floor} as  \textit{standing on},  and also misses the spatial relationships such as  \textit{behind},  \textit{left} between \textit{garbage} and \textit{object}. In contrast, our method correctly recovers the  \textit{lying on} relation and surrounding spatial context (e.g., \textit{close by}, \textit{front, right}) through relational reasoning.

In the second example with three  \textit{chairs} arranged symmetrically around a table. 3DHetSGP incorrectly predicts asymmetric  relations such as \textit{higher than} / \textit{right} and \textit{lower than} / \textit{left}, even though all chairs are semantically and geometrically identical.  This demonstrates its limitation in modeling interdependent spatial relations. Our method accurately  predicts \textit{same as} relations and consistent spatial relations among all the chairs, demonstrating a stronger ability to model context consistency.

In the third example, 3DHetSGP fails to predict the \textit{standing on} relation between \textit{plant} and \textit{cabinet}, predicting \textit{none}. Our method successfully restores physically and spatially consistent relations.

These examples collectively highlight the limitations of object-centric models in resolving relational ambiguity and spatial correlation. In contrast, our approach enhances relational reasoning through explicit modeling of dependencies among relations, resulting in more complete and semantically accurate scene graphs.

\begin{figure*}[!t]
\centering
\resizebox{0.95\textwidth}{!}{\includegraphics[width=4.5in]{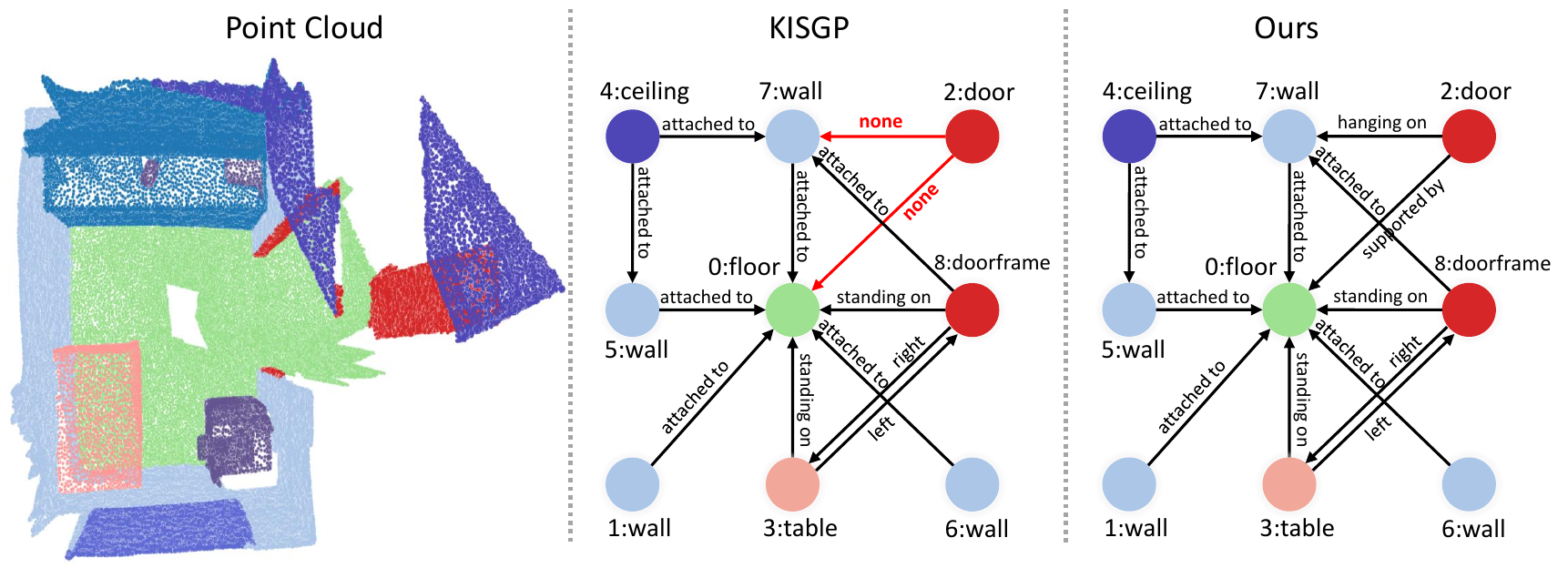}} \\
\caption{Qualitative results of our model and baseline KISGP \cite{KISGP} on ScanNet.  Red arrows indicate missed predicted relationships.  }
\label{Q_SN}
\end{figure*}

\noindent  \textbf {B2. Qualitative Analysis on ScanNet dataset. }
In Figure \ref{Q_SN}, presents a qualitative comparison between our method with KISGP on a ScanNet scene. 
KISGP predicts a considerable set of relations in this ScanNet scene. Our method predicts a larger set and identifies additional structural and spatial interactions, including \textit{supported by} between the \textit{door} and the \textit{floor} and  \textit{hanging on} between the \textit{door} and the \textit{wall}. These additional relations produce a more complete and more coherent scene graph,  further demonstrating the robustness of our relational reasoning.

\end{document}